\documentclass[letterpaper]{article} 
\usepackage{aaai25}  
\usepackage{times}  
\usepackage{helvet}  
\usepackage{courier}  
\usepackage[hyphens]{url}  
\usepackage{graphicx} 
\usepackage{natbib}  
\usepackage{caption} 
\usepackage{algorithm}
\usepackage{algorithmic}
\usepackage{amsmath}
\usepackage{amssymb}
\usepackage{dsfont}
\usepackage{booktabs}
\usepackage{multirow}
\usepackage{newfloat}
\usepackage{listings}
\usepackage{mathtools}
\usepackage{amsthm}
\newcommand{\indicator}[1]{\mathds{1}{#1}}

\usepackage{newfloat}
\usepackage{listings}
\DeclareCaptionStyle{ruled}{labelfont=normalfont,labelsep=colon,strut=off} 
\floatstyle{ruled}
\newfloat{listing}{tb}{lst}{}
\floatname{listing}{Listing}
\title{Personalized Clustering via Targeted Representation Learning}
\author {
    Xiwen Geng\textsuperscript{\rm 1,\rm 2},
    Suyun Zhao\textsuperscript{\rm 1,\rm 2}\thanks{Corresponding authors.},
    Yixin Yu\textsuperscript{\rm 3},
    Borui Peng\textsuperscript{\rm 3},
    Pan Du\textsuperscript{\rm 1,\rm 2}\\
    Hong Chen\textsuperscript{\rm 1,\rm 2},
    Cuiping Li\textsuperscript{\rm 1,\rm 2},
    Mengdie Wang\textsuperscript{\rm 1,\rm 2}
}
\affiliations {
    \textsuperscript{\rm 1}Key Lab of Data Engineering and Knowledge Engineering of MOE Renmin University of China\\
    \textsuperscript{\rm 2}School of Information, Renmin University of China\\
    \textsuperscript{\rm 3}School of Statistics, Remin University of China\\
    \{xiwen, zhaosuyun, yuyixin3642, pengborui, chong, licuiping\}@ruc.edu.cn, \{du\_pan, wangmd\_ruc\}@163.com
}

\usepackage{bibentry}

\begin{document}

\maketitle

\begin{abstract}
Clustering traditionally aims to reveal a natural grouping structure within unlabeled data. However, this structure may not always align with users' preferences. In this paper, we propose a personalized clustering method that explicitly performs targeted representation learning by interacting with users via modicum task information (e.g., \textit{must-link} or \textit{cannot-link} pairs) to guide the clustering direction. We query users with the most informative pairs, i.e., those pairs most hard to cluster and those most easy to miscluster, to facilitate the representation learning in terms of the clustering preference. Moreover, by exploiting attention mechanism, the targeted representation is learned and augmented. By leveraging the targeted representation and constrained contrastive loss as well, personalized clustering is obtained. Theoretically, we verify that the risk of personalized clustering is tightly bounded, guaranteeing that active queries to users do mitigate the clustering risk. Experimentally, extensive results show that our method performs well across different clustering tasks and datasets, even when only a limited number of queries are available.
\end{abstract}

\section{Introduction}
Deep clustering refers to some unsupervised machine learning techniques that leverage deep learning to reveal the grouping structures hidden in data samples. Unlike traditional clustering methods, deep clustering aims to learn latent representations of the data that facilitate clustering in terms of its underlying patterns
~\cite{Xie2016UnsupervisedDE,Caron2018DeepCF,Li2021ContrastiveC,Zhong2021GraphCC,Liu2022DeepGC,li2022twin,li2023image}. 
\begin{figure}[t]
\centering
\includegraphics[width=0.9\columnwidth]{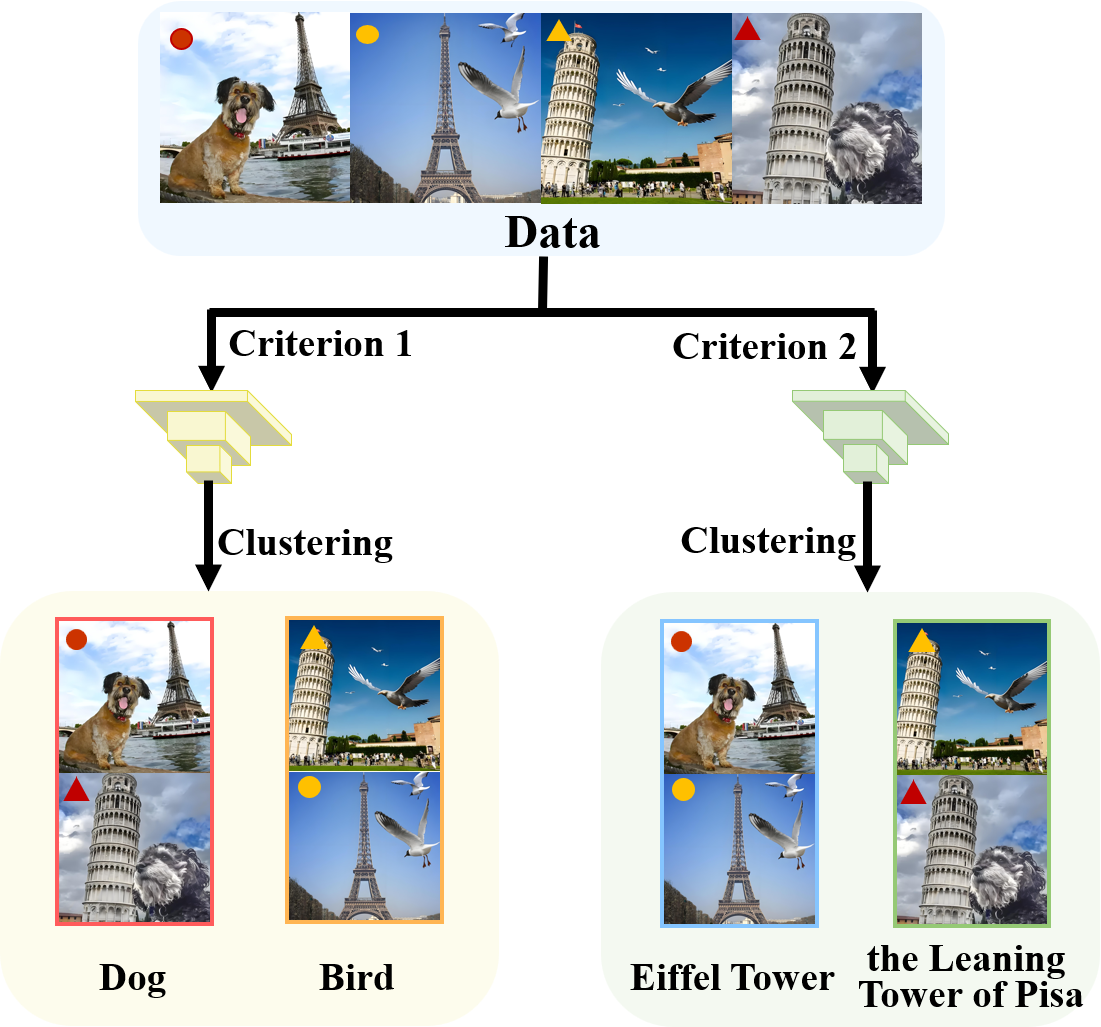} 
\caption{The diversity of cluster orientation. Different tasks have different orientations for feature learning and image clustering.}
\label{fig1}
\end{figure}

Usually, most of the existing methods conduct clustering with the unique goal of maximizing clustering performance, while ignoring the personalized demand of clustering tasks. In real scenarios, however, users may tend to cluster unlabeled data according to their preferences, such as distinctive objectives (animals, architectures, and characters etc.), and then some personalized clustering demands are put forward. For example, Fig.1 depicts the diverse criteria of clustering. There are multiple objects in each image, such as animals and architectures. Some users require clustering in terms of animals, so the two images with dogs should be grouped in the same cluster, classified as dogs. In contrast, some others may prefer to group images according to architectures; thus, the two images with the Eiffel Tower should be clustered together. In cases with such personalized demands, many existing clustering techniques may decline or even be unworkable without user guidance.Therefore, it is still a challenging problem to cluster along a desired orientation.

A candidate solution to this issue is constrained clustering, which incorporates prior knowledge through constraints to guide the neural network toward a desired cluster orientation~\cite{Wagstaff2001ConstrainedKC}. The most commonly used constraint is the pairwise constraint, which provides information by indicating whether a pair of samples belongs to the same (\textit{must-link}) or distinct (\textit{cannot-link}) clusters. While they use constraints to facilitate models to achieve superior clustering performance, rather than towards a desired configuration. Sometimes, randomly selected constraint pairs~\cite{Manduchi2021DeepCG,Sun2022ActiveDI}  may lead to inferior clustering results in personalized scenarios, just as demonstrated in the experiments in Section 4. Accordingly, actively querying informative sample pairs may facilitate personalized clustering. Unfortunately, 
several proposed active query strategies~\cite{Sun2022ActiveDI,Pei2015BayesianAC,Biswas2014ActiveIC} tend to pick those samples beneficial for the default cluster orientation~\cite{Sun2022ActiveDI}.
Typically, their learned representations do not align with the desired cluster orientation, resulting in poor performance on personalized clustering~\cite{Angell2022InteractiveCC}.

To narrow this gap, we propose the personalized clustering model via targeted representation learning (PCL), which selects the most valuable sample pairs to learn targeted representations, thereby achieving clustering in the desired orientation. Specifically, we propose an active query strategy which picks those pairs most hard to cluster and those most easy to mis-cluster, to facilitate the representation learning in terms of the clustering task. Simultaneously, a constraint loss is designed to control targeted representation learning in terms of the cluster orientation. By exploiting attention mechanism, the targeted representation is augmented. Extensive experiments demonstrate that our method is effective in performing personalized clustering tasks. Furthermore, by strict mathematical reasoning, we verify the effectiveness of the proposed PCL method. 

To summarize, our main contributions are listed as follows:
\begin{itemize}
    \item We propose a novel model, called PCL, for a personalized clustering task, which leverages active query to control the targeted representation learning.
    \item A theoretical analysis on clustering is conducted to verify the effectiveness of active query in constraint clustering. Simultaneously, the tight upper bound of generalization risk of PCL is given.
    \item Extensive experiments show that our model outperforms five unsupervised clustering approaches, four semi-supervised clustering approaches on three image datasets. Our active query strategy also performs well when compared to other methods.
\end{itemize}

The remainder of this paper is organized as follows. Related works are reviewed in Section \ref{sec:2}. We propose our method in Section \ref{sec:3}. Furtherly, Section \ref{sec:4} evaluated our proposed approach experimentally. We conclude our work in Section \ref{sec:5}.

\begin{figure*}[ht]
\centering
\includegraphics[width=0.95\textwidth]{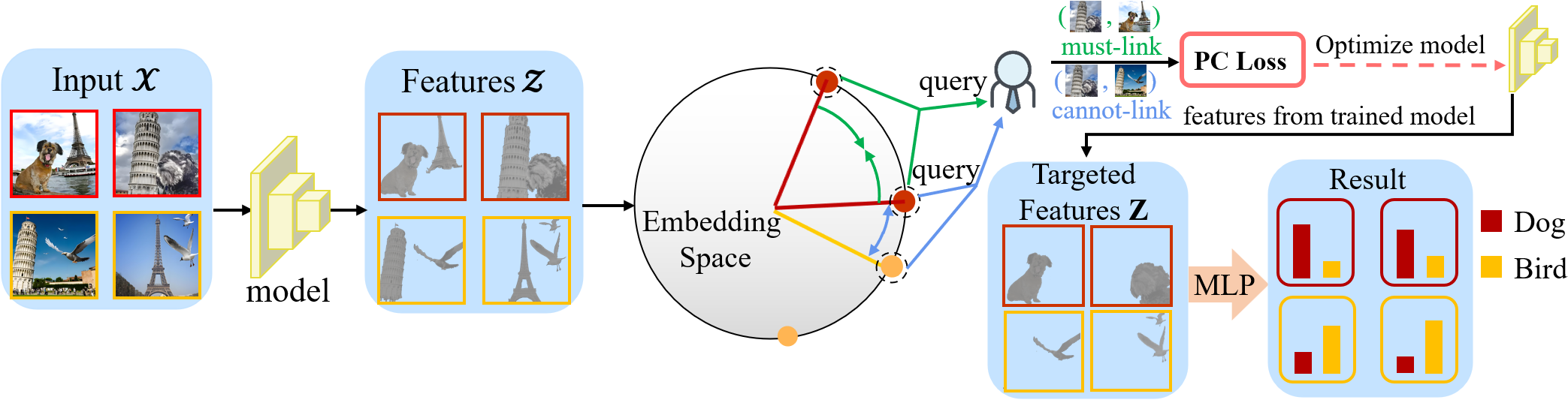} 
\caption{The framework of PCL. A deep neural network creates representations from two random augmentations of the data. By assessing the position of images in the feature space, PCL selects the most informative sample pairs to guide the training of the model. The model is then retrained by querying whether these pairs are \textit{must-link} or \textit{cannot-link}. This approach allows the final model to concentrate on features relevant to the desired cluster orientation, resulting in accurate clustering outcomes.}
\label{fig2}
\end{figure*}

\section{Related Work}
\label{sec:2}
\subsection{Deep Clustering}
Deep neural networks have been explored to enhance clustering performance due to their powerful ability of representation learning ~\cite{Vincent2010StackedDA,Kingma2014AutoEncodingVB}. A notable method is DEC~\cite{Xie2016UnsupervisedDE}, which optimizes cluster centers and features simultaneously by minimizing the KL-divergence in the latent subspace. Similarly, IDFD~\cite{TaoTN21} improves data similarity learning through sample discrimination and then reduces redundant correlations among features via feature decorrelation.

Recently, more and more methods achieve clustering representations by leveraging contrastive learning techniques, thereby facilitating deep clustering. For instances, DCCM~\cite{Wu2019DeepCC} mines various kinds of sample relations by contrastive learning techniques and then train its network using the highly-confident information. PICA~\cite{Huang2020DeepSC} minimizes the cosine similarity between cluster-wise assignment vectors to achieve the most semantically plausible clustering solution. Moreover, CC~\cite{Li2021ContrastiveC} proposes a dual contrastive learning framework aimed at achieving deep clustering representations. GCC~\cite{Zhong2021GraphCC} selects positive and negative pairs using a KNN graph constructed on sample representations. Meanwhile, SPICE~\cite{niu2022spice} divides the clustering network into a feature model and a clustering head, splitting the training process into three stages. 

Though these deep clustering methods perform well in default orientation, they work poorly in personalized tasks. 

\subsection{Constrained Clustering}
Constrained clustering refers to a type of clustering that incorporates additional information into the clustering process, e.g., pairwise (\textit{must-link} and \textit{cannot-link}) constraints~\cite{Chien2019HS2AL}.

Cop-Kmeans~\cite{Wagstaff2001ConstrainedKC} was the first to introduce pairwise constraints to traditional K-means clustering algorithms to enhance clustering performance, and many subsequent works are then proposed ~\cite{Basu2002SemisupervisedCB,bilenko2004integrating,Zheng2011SemisupervisedHC,Yang2012ConsensusCB,Yang2013AnIC}. In recent years, constrained clustering approaches have been combined with deep neural networks. For instance, by introducing pairwise constraints, SDEC~\cite{Ren2019SemisupervisedDE} extends the work of DEC~\cite{Xie2016UnsupervisedDE}, and DCC~\cite{Zhang2019AFF} extends the work of IDEC~\cite{Guo2017ImprovedDE}. Bai et al. explored ways to improve clustering quality in scenarios where constraints are sourced from different origins~\cite{bai2020semi}. DC-GMM~\cite{Manduchi2021DeepCG} explicitly integrates domain knowledge in a probabilistic form to guide the clustering algorithm toward a data distribution that aligns with prior clustering preferences. CDAC+~\cite{Lin2019DiscoveringNI} constructs pair constraints to transfer relations among data and discover new intents (clusters) in dialogue systems. Pairwise constraints are widely used to direct clustering models.

Some constrained clustering methods also integrate active learning~\cite{Ren2022ASO}, which aims to query the most informative samples to reduce labeling costs while maintaining performance~\cite{Ashtiani2016ClusteringWS}. COBRA~\cite{Craenendonck2018COBRAAF} merges small clusters resulting from K-means into larger ones based on pairwise constraints by maximally exploiting constraint transitivity and entailment. ADC~\cite{Sun2022ActiveDI} constructs contrastive loss by comparing pairs of constraints and integrates deep representation learning, clustering, and data selection into a unified framework. 

Although these methods typically cluster according to prior preferences, they do not consider the diverse criteria of clustering but focus on improving performance or reducing the query budget. Dasgupta and Ng proposed an active spectral clustering algorithm that specifies the dimension along the sentiment embedded in the data points, which is similar to our problem of diverse image clustering~\cite{dasgupta2010clustering}. However, their method relies on label constraints and is designed for natural language processing.

\section{Methods}
\label{sec:3}
\subsection{Personlized Clustering}
\label{sec:3-1}
We propose a novel Personalized Clustering model (PCL) that leverages active learning to guide the model's clustering in a specified orientation. Using a designed pairwise scoring function within a query strategy, certain image pairs are selected for annotation. Oracles then judge whether these image pairs belong to the same class, responding with \textit{YES} or \textit{NO}. These responses form \textit{must-link} or \textit{cannot-link} constraints, which are incorporated into the construction of a constrained contrastive loss, adjusting the network parameters through backpropagation. By integrating the cross-instance attention module in this way, the model can capture features that align more closely with the desired cluster orientation~\cite{gan2023superclass}. The specific framework of PCL is illustrated in Figure 2.

Given a mini-batch of images $\mathcal{X}=\{\mathbf {x}_1,...,\mathbf{x}_N\}$ where $N$ is the batch size, we apply two stochastic data transformations, $T^a$ and $T^b$, from the same family of augmentations $\mathcal T$ , resulting in augmented data $\{\mathbf x_1,...,\mathbf x_N,\mathbf x_{N+1},...,\mathbf x_{2N}\}$. This process constructs the initial positive pairs $(\mathbf x_i,\mathbf x_{i+N})_{i=1}^N$. The network then extracts preliminary features from the augmented data and refines them using the cross-instance attention module. It projects these representations into the feature space, yielding target features $\{\mathbf z_1,...,\mathbf z_{2N}\}$, and into the cluster space, yielding assignment probabilities $\{\mathbf p_1,...,\mathbf p_{2N}\}$. These are expressed in matrices $\mathbf Z\in\mathbb{R}^{2N\times M}$ and $\mathbf P\in\mathbb{R}^{2N\times K}$, where $M$ is the feature dimension and $K$ is the number of clusters. 

Next, we select the most informative pair based on the scoring function, submit a query to the oracle, and document the feedback in the indicator matrix $\indicator{}^+$ and $\indicator{}^-$. These matrices represent the relationship of pairs, where $\indicator{}^+_{ij}=1$ if $y_i=y_j$(\textit{must-link}) and $\indicator{}^-_{ij}=1$ if $y_i\neq y_j$(\textit{cannot-link}). Meanwile, due to the initial positive pairs, we use the indicator matrix $\indicator{}$ to represent the positive pair $(\mathbf{x}_{i}$, $\mathbf{x}_{j})$ is constructed by data augment, where $\indicator{}_{ij}=1$ if $(\mathbf x_i,\mathbf x_j)$ is constructed by data augment. These constraints guide the clustering process to the desired orientation via the constrained contrastive loss.

In the following two subsections, we first introduce our constrained contrastive loss in Subsection \ref{sec:3-1}, followed by the pairwise query strategy in Subsection \ref{sec:3-2}.

\subsection{Constrained Contrastive Loss}
\label{sec:3-2}
To achieve effective and targeted clustering, we employ a constrained contrastive loss similar to infoNCE~\cite{oord2018representation} to evaluate the loss of each sample. Different weights are set for positive pairs based on their construction methods. We focus on clustering impacts in both feature and clustering spaces, while also aiming to differentiate category representations in the clustering space. In the feature space, the loss function for a positive pair of examples$(i,j)$ is defined as:
\begin{equation}
\label{eq:1}
    l(\mathbf z_i,\mathbf z_j,\mathbf Z)=-\log\frac{\exp(s(\mathbf z_i,\mathbf z_j)/\tau)}{\mu_i\sum\limits_{k=1}\limits^{2N}(1+\indicator{}_{ik}^-)\exp(s(\mathbf z_i,\mathbf z_k)/\tau)},
\end{equation}
where $s(\mathbf a, \mathbf b)=(\mathbf a\mathbf b^\top)/(\lVert \mathbf a\rVert \lVert \mathbf b\rVert)$ measures pair-wise similarity by cosine distance, $\mu_i=N/(N+\sum_{j=1}^{2N}(\indicator{}^-_{ij}))$ normalizes the range of the denominator, and $\tau$ is the temperature parameter.

Positive pairs from data augmentation and queries are weighted differently in the loss term. A symmetric matrix $\textbf W \in \mathcal R^{2N\times 2N}$ is constructed with elements:
\begin{equation}
\label{eq:2}
\left.w_{ij}:=\left\{
\begin{array}{ll}
N_+,&\text{if } \indicator{}_{ij}=1\\
\lambda N,&\text{if } \indicator{}_{ij}^+=1\\
0,&\text{otherwise}
\end{array}
\right.\right.
\end{equation}
where $\lambda$ is the hyperparameter and $N_+$ is the total number of positive pairs constructed by querying.

The loss of sample $i$ is defined as:
\begin{equation}
\label{eq:3}
    \ell_i=\sum_{j=1}^{2N}\frac{w_{ij}(l(\mathbf z_i,\mathbf z_j,\mathbf Z)+l(\mathbf p_i,\mathbf p_j,\mathbf P))}{\sum_{k=1}^{2N}w_{ik}}.
\end{equation}

In the clustering space, it is crucial to ensure that each cluster remains distinct. Letting $\mathbf C=\mathbf P^\top\in \mathcal{R}^{K\times 2N}$, we obtain a set $\{\mathbf c_1,...,\mathbf c_K,\mathbf c_{K+1},...,\mathbf c_{2K}\}$, where $\mathbf c_i (i\leq K)$ and $\mathbf c_{i+K}$ represent the i-th cluster under augmentations $T^a(\mathcal X)$ and $T^b(\mathcal X)$. A loss function is defined to differetiate i-th cluster from others:
\begin{equation}
\label{eq:4}
    \hat \ell_i=l(\mathbf c_i,\mathbf c_{i+N}, \mathbf C)+l(\mathbf c_{i+N},\mathbf c_{i}, \mathbf C)
\end{equation}

Our overall loss is defined as:
\begin{equation}
\label{eq:5}
\mathcal L_{PC}=\frac{1}{2N}\sum_{i=1}^{2N}\ell_i+\frac{1}{2K}\sum_{i=1}^{K}\hat \ell_i+H(\hat Y)
\end{equation}
where $ H(\hat Y)=-\sum_{i=1}^K[P(\hat Y_i)log(P(\hat Y_i))]$ serves as a regularization term.

\begin{figure}[t]
\centering
\includegraphics[width=1\columnwidth]{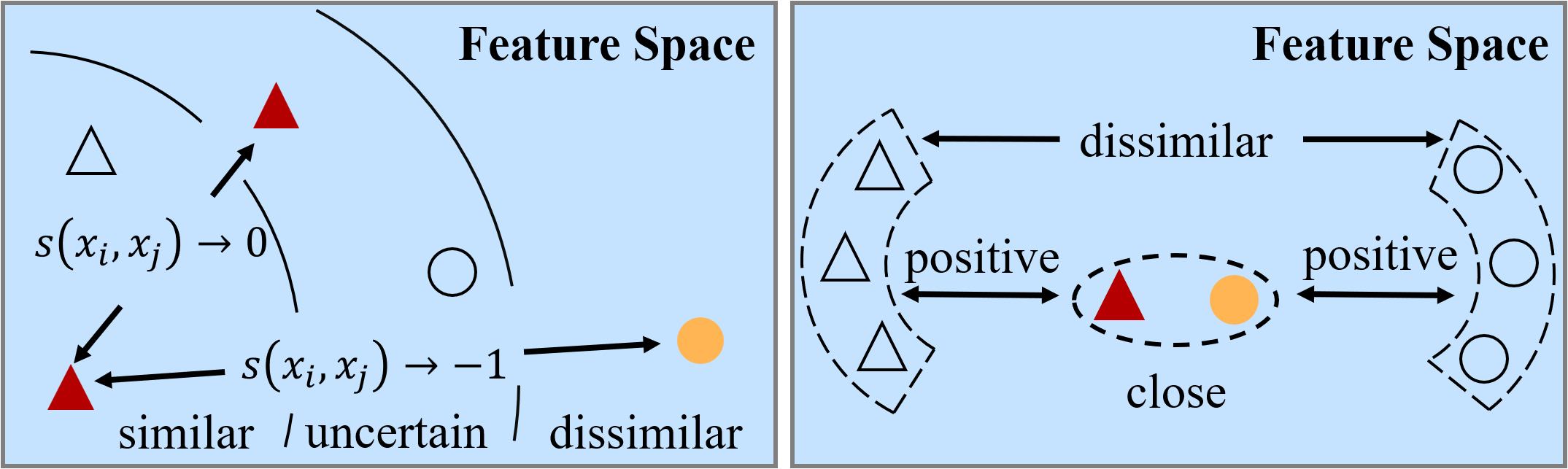}
\caption{(Left) The illustration shows a sample pair with uncertain similarity, highlighted as red nodes, which are prioritized for querying. (Right) Some samples, due to incorrect feature extraction, are mistakenly grouped together despite belonging to different clusters, as shown by the red and yellow nodes. Identifying these can be achieved by assessing the similarity between their positive samples.}
\label{fig3}
\end{figure}

\subsection{Query Strategy}
\label{sec:3-3}
Our model learns features conducive to clustering by utilizing a constrained contrastive loss, emphasizing the importance of selecting informative sample pairs to guide cluster orientation. We propose an innovative query strategy that considers both the uncertainty of sample pairs and hard negatives, as illustrated in Fig.\ref{fig3}. This strategy identifies critical sample pairs to refine cluster orientation and enhance performance.

Since features extracted by untrained models are initially unreliable, random constraints are generated on the data before training, and the model is pre-trained before applying the query strategy. We believe that the initial aim is to establish a general cluster orientation and it becomes essential to rectify samples misclassified by the model and make minor adjustments to the cluster orientation later. Therefore, at the outset of the task, the uncertainty of sample pairs takes precedence. As the task progresses, focusing on hard negatives becomes crucial.

\textbf{Uncertainty of pairs.}
To establish cluster orientation, we prioritize the selection of the most uncertain sample pairs. Cosine similarity of sample pairs ranges from $-1$ to $1$. Pairs with similarity near $1$ likely belong to the same cluster, whereas pairs near $-1$ are likely in different clusters. Pairs with similarity close to $0$ are uncertain for the model, offering both similar and distinct features that provide valuable insights into cluster orientation. The uncertainty score for an sample pair $(\mathbf x_i,\mathbf x_j)$ is defined as:

\begin{equation}\label{eq:6}
\mathcal S_{up}(\mathbf x_i,\mathbf x_j)=\sigma[-|s(\mathbf z_i,\mathbf z_j)-\epsilon|],
\end{equation}

where $\mathbf z_i$ and $\mathbf z_j$ are the features of samples $\mathbf x_i$ and $\mathbf x_j$ in feature space. Similarity is averaged over two data augmentations. $\sigma[\cdot]$ denotes the Min-Max normalization operator~\cite{jain2005score}, and $\epsilon$ is a small positive number as \textit{must-link} constraints are more informative than \textit{cannot-links}~\cite{Sun2022ActiveDI}. 

\textbf{Hard negatives.}
Hard negatives are samples with a significant discrepancy between the actual clustering and the current results of the model. Although actual clustering results cannot be directly observed, they can be inferred using known positive and negative examples~\cite{zhuang2023deep}. For two samples $\mathbf x_i$ and $\mathbf x_j$, $P(\mathbf x_i,\mathbf x_j)=\sigma[s(\mathbf x_i, \mathbf x_j)]$ represents the probability that these samples are in the same cluster. $Q(\mathbf x_i,\mathbf 
x_j)=\sigma[\sum_{k=1}^N(\indicator{}_{jk}^+s(\mathbf x_i,\mathbf x_k)-\indicator{}_{jk}^-s(\mathbf x_i,\mathbf x_k))]$ represents the probability calculated from the known pairs. Relative entropy is used to assess the difference between $P(\mathbf x_i,\mathbf x_j)$ and $Q(\mathbf x_i,\mathbf x_j)$, identifying challenging samples to classify:
\begin{equation}\label{eq:7}
\mathcal S_{hp}(\mathbf x_i,\mathbf x_j)=\sigma[P(\mathbf x_i, \mathbf x_j)\log\frac{P( \mathbf x_i,\mathbf x_j)}{Q(\mathbf x_i,\mathbf x_j)}].
\end{equation}

\textbf{Query strategy.}
The strategy for querying the most informative sample pairs combines the scores defined above. The joint query score is given by:
\begin{equation}\label{eq:8}
\mathcal S(\mathbf x_i,\mathbf x_j)= r[\mathcal S_{up}(\mathbf x_i,\mathbf x_j)]+(1-r)\mathcal S_{hp}(\mathbf x_i,\mathbf x_j)
\end{equation}

where $r$ is the ratio of the remaining query budget to the total budget, indicating that the weight of hard-to-learn pairs increases with more queries. As queries increase, the cluster orientation is refined and the strategy focuses on maximizing performance.

The full querying and clustering algorithm\footnote{Code is available at https://github.com/hhdxwen/PCL.} of the model is detailed in the Appendix.\footnote{Appendix is available at https://arxiv.org/abs/2412.13690.}

\subsection{Theoretical Analysis}
\label{sec:3-4}
We focus on the feature extractor function  $f \in \mathcal{F}: \mathcal{X} \rightarrow Z$ which determines the final clustering results \cite{pmlr-v97-saunshi19a}. The unsupervised loss of sample $i$ is defined as: 
\begin{equation}
\label{eq:9}
    \ell_i^{'}(f)=\sum_{j=1}^{2N}\frac{w_{ij}(l(f(\mathbf x_i),f(\mathbf x_j),\mathbf Z)}{\sum_{k=1}^{2N}w_{ik}}.
\end{equation}
\subsubsection{Definition 3.1 (Unsupervised Loss)}\textit{ If we can query $q$ times, the overall unsupervised loss is defined as:}
\begin{equation}
\label{eq:10}
    \hat{\mathcal{L}}_q(f) \coloneqq \frac{1}{2N}\sum_{i=1}^{2N}\ell_i^{'}(f), \quad
    q \coloneqq \sum_{1\leq i < j \leq 2N} (\indicator{}_{ij}^{+} + \indicator{}_{ij}^{-}).
\end{equation}
\textit{and $\hat{f}_{q}$ is defined as $\arg \min_{f \in \mathcal{F}} \hat{\mathcal{L}}_{q}(f)$. }

\textit{The unsupervised population loss after $q$ queries is defined as:}
\begin{equation}
\label{eq:11}
    \mathcal{L}_q(f) \coloneqq \mathbb{E}_{x \sim \mathcal{X}}[\hat{\mathcal{L}}_q(f)],
\end{equation}
\textit{and $f^{*}_{q}$ is defined as $\arg \min_{f \in \mathcal{F}} \mathcal{L}_{q}(f)$.}

Firstly we propose Theorem 3.1. All proofs are in Appendix.

\subsubsection{Theorem 3.1} \textit{Suppose we can completely query each sample. We use $Q$ as the total number of queries. With probability at least $1-\delta$},
\begin{equation}
\label{eq:13}
    | \mathcal{\hat{L}}_{Q}(\hat{f}_{Q}) - \mathcal{L}_{Q}(f^{*}_{Q}) | \leq \mathcal{O}(\frac{\eta \mathcal{R}_{\mathcal{S}}(\mathcal{F})}{N} + \sqrt{\frac{\log\frac{1}{\delta}}{N}}).
\end{equation}
\textit{where $\mathcal{R}_{\mathcal{S}}(\mathcal{F})$ is the Rademacher complexity of $\mathcal{F}$ with respect to training set $\mathcal{S}$, and $\eta$ is the Lipschitz constant.}

Theorem 3.1 indicates that $\mathcal{L}_{Q}(f^{*}_{Q})$ can be well bounded if the trained model $\hat{f}_{Q}$ is good.
However, due to the limited queries (less than $Q$), we cannot calculate $\mathcal{\hat{L}}_{Q}(\hat{f}_{Q})$. As a compromise, can we get closer to it?

\subsubsection{Theorem 3.2 (Queries Reduce the Gap).} \textit{We only consider adding one query. If $q \ll N$:}
\begin{equation}
\label{eq:14}
    |\mathcal{\hat{L}}_{q}(\hat{f}_{q}) - \mathcal{\hat{L}}_{Q}(\hat{f}_{Q})| \geq |\mathcal{\hat{L}}_{q+1}(\hat{f}_{q+1}) - \mathcal{\hat{L}}_{Q}(\hat{f}_{Q})|.
\end{equation}

This just matches the process of algorithm: 
we initially make a new query request based on the existing model $\hat{f}_{q}$, 
followed by updating $\mathcal{\hat{L}}_{q}$ to $\mathcal{\hat{L}}_{q+1}$ based on the query results, 
and finally optimizing the loss to get $\hat{f}_{q+1}$. 
Theorem 3.2 guarantees the gap between the loss of the current model and the complete queries loss will be reduced after each query. Now we show that our strategy is indeed effective.

\subsubsection{Theorem 3.3 (Query Strategy Is Effective).}\textit{Suppose $q \ll N$, the sample pair $(\mathbf x_i, \mathbf x_j)$ is queried, and the result is cannot-link,  we have:}
\begin{equation}
\label{eq:15}
    |\mathcal{\hat{L}}_{q}(f) - \mathcal{\hat{L}}_{q+1}(f)| \propto  \mathcal \exp\left(\mathcal{S}_{up}(\mathbf x_i,\mathbf x_j)\right),
\end{equation}
\textit{and if the result is must-link, we have:}
\begin{equation}
\label{eq:16}
    |\mathcal{\hat{L}}_{q}(f) - \mathcal{\hat{L}}_{q+1}(f)| \propto \mathcal S_{up}(\mathbf x_i,\mathbf x_j).
\end{equation}

Since we want to reduce the gap in Theorem 3.2, we should equivalently maximize $|\mathcal{\hat{L}}_{q}(f) - \mathcal{\hat{L}}_{q+1}(f)|$ and then train the model according to $\mathcal{\hat{L}}_{q+1}$.
Theorem 3.3 states that our query strategy is effective in this sense, 
because when $q$ is small, $\mathcal{S}_{up}$ matters more in the scoring function.
Besides, we have added $\mathcal{S}_{hp}$ in the score and weight it more if $q$ is larger, making our strategy more robust and efficient. 
The following experiments have proved the effectiveness in practice.

\section{Experiments}
\label{sec:4}
In this section, we present a comprehensive set of experiments to evaluate the effectiveness of the proposed method.
\begin{table}[t]   
\begin{center}
\begin{tabular}{cccc}   
\toprule  
\textbf{Dataset} & \textbf{Instances} & \textbf{Classes} & \textbf{Clusters} \\   
\midrule   
CIFAR10-2 & 60,000 & 10 & 2  \\   
CIFAR100-4 & 60,000 & 100 & 4  \\   
ImageNet10-2  & 12,818 & 10 & 2  \\   
\bottomrule   
\end{tabular}   
\caption{A summary of the datasets.}  
\label{table:1} 
\end{center}   
\end{table}
\begin{table*}[ht]
  \centering
    \resizebox{\textwidth}{!}{
    \begin{tabular}{cccccccccccccc}
    \toprule
          &       & \multicolumn{4}{c}{CIFAR10-2} & \multicolumn{4}{c}{CIFAR100-4} & \multicolumn{4}{c}{ImageNet10-2}  \\
    \midrule
    \multicolumn{2}{c}{Default} & NMI   & ARI & F  & ACC   & NMI   & ARI & F  & ACC   & NMI   & ARI  &  F  & ACC    \\
    \midrule
    \multirow{5}[2]{*}{Unsupervised} 
          & kmeans & 0.054  & 0.079  & 0.550  & 0.641  & 0.038  & 0.033  & 0.281  & 0.358  & 0.078  & 0.103 & 0.553 & 0.661  \\
          & MiCE  & 0.664  & 0.755  & 0.881  & 0.934  & 0.230  & 0.228  & 0.423  & 0.590  & 0.762  & 0.849 & 0.925 & 0.960  \\
          & CC   & 0.575  & 0.662  & 0.835  & 0.907  & 0.176  & 0.166  & 0.379  & 0.515  & 0.508  & 0.615 & 0.807 & 0.892  \\
          & GCC    & 0.653  & 0.753  & 0.880  & 0.934  & \textbf{0.344}  & \textbf{0.337}  & \textbf{0.503}  & \textbf{0.604}  & \textbf{0.908}  & \textbf{0.952} & \textbf{0.976} & \textbf{0.988}  \\
          & SPICE  & 0.600  & 0.638  & 0.823  & 0.899  & 0.250  & 0.197  & 0.399 & 0.496 & 0.642 & 0.725 & 0.864 & 0.925  \\
    \midrule
    \multirow{5}[4]{*}{Constrained} 
          & DCC    & 0.572  & 0.686  & 0.848  & 0.914  & 0.134  & 0.120  & 0.345  & 0.449  & 0.432  & 0.536 & 0.768 & 0.866  \\
          & ADC    & 0.544  & 0.662  & 0.838  & 0.907  & 0.047  & 0.055  & 0.319  & 0.390  & 0.453  & 0.550 & 0.775 & 0.870  \\
          & SDEC   & 0.355  & 0.430  & 0.720  & 0.828  & 0.070  & 0.048  & 0.317  & 0.380  & 0.192  & 0.254 & 0.627 & 0.752  \\
          & DC-GMM & 0.542  & 0.657  & 0.838  & 0.905  & 0.107  & 0.121  & 0.349  & 0.431  & 0.490  & 0.598 & 0.799 & 0.886 \\
\cmidrule{2-14}          & Ours   & \textbf{0.737} & \textbf{0.830} & \textbf{0.918} & \textbf{0.955} & 0.197 & 0.208 & 0.407 & 0.575 & 0.722 & 0.817 & 0.908 & 0.952 \\
    \bottomrule
    \end{tabular}%
    }
  \caption{The clustering performance under the default orientation, on three object image benchmarks. The default orientation completely follows the tendency of deep clustering. The best results are shown in boldface.}
  \label{table:2-1}
\end{table*}%

\begin{table*}[ht]
  \centering
    \resizebox{\textwidth}{!}{
    \begin{tabular}{cccccccccccccc}
    \toprule
          &       & \multicolumn{4}{c}{CIFAR10-2} & \multicolumn{4}{c}{CIFAR100-4} & \multicolumn{4}{c}{ImageNet10-2}  \\
    \midrule
    \multicolumn{2}{c}{Personlized} & NMI   & ARI & F  & ACC   & NMI   & ARI & F  & ACC   & NMI   & ARI  &  F  & ACC    \\
    \midrule
    \multirow{5}[2]{*}{Unsupervised} 
          & kmeans & 0.009  & 0.015  & 0.519  & 0.563  & 0.028  & 0.022  & 0.274  & 0.314  & 0.027  & 0.037 & 0.519 & 0.597  \\
          & MiCE   & 0.028  & 0.045  & 0.535  & 0.607  & 0.159  & 0.135  & 0.355  & 0.399  & 0.024  & 0.032 & 0.516 & 0.591  \\
          & CC     & 0.022  & 0.035  & 0.528  & 0.594  & 0.132  & 0.115  & 0.342  & 0.405  & 0.032  & 0.043 & 0.521 & 0.605  \\
          & GCC    & 0.035  & 0.057  & 0.542  & 0.620  & 0.230  & 0.203  & 0.404  & 0.508  & 0.025  & 0.034 & 0.517 & 0.592  \\
          & SPICE  & 0.005  & 0.007  & 0.513  & 0.543  & 0.232  & 0.215  & 0.413  & 0.504  & 0.068  & 0.088 & 0.548 & 0.648  \\
    \midrule
    \multirow{5}[4]{*}{Constrained} 
          & DCC    & 0.244  & 0.334  & 0.679  & 0.789  & 0.099  & 0.095  & 0.340  & 0.473  & 0.204  & 0.267 & 0.634 & 0.758  \\
          & ADC    & 0.206  & 0.291  & 0.665  & 0.771  & 0.024  & 0.023  & 0.374  & 0.322  & 0.102  & 0.138 & 0.570 & 0.686  \\
          & SDEC   & 0.013  & 0.021  & 0.521  & 0.573  & 0.092  & 0.083  & 0.354  & 0.417  & 0.075  & 0.102 & 0.552 & 0.660  \\
          & DC-GMM & 0.060  & 0.074  & 0.671  & 0.655  & 0.072  & 0.060  & 0.311  & 0.384  & 0.000  & 0.000 & 0.675 & 0.512  \\
\cmidrule{2-14}          & Ours   & \textbf{0.453} & \textbf{0.558} & \textbf{0.765} & \textbf{0.873} & \textbf{0.234} & \textbf{0.247} & \textbf{0.437} & \textbf{0.597} & \textbf{0.409} & \textbf{0.510} & \textbf{0.755} & \textbf{0.857} \\
    \bottomrule
    \end{tabular}%
    }
  \caption{The clustering performance under the personalized orientation, on three object image benchmarks. The personalized orientation is artificially designed opposite to the default one. The best results are shown in boldface.}
  \label{table:2-2}
\end{table*}%
\subsection{Experiments Settings}
\label{sec:4-1}
\textbf{Datasets.} We assessed our method using three widely-used image datasets: CIFAR-10, CIFAR-100~\cite{Krizhevsky2009LearningML}, and ImageNet-10~\cite{deng2009imagenet}. To demonstrate that constraints can guide cluster orientation, we differentiate between training and test sets, applying constraints only to the training set. For CIFAR-10, CIFAR-100, and ImageNet-10, we added 10k, 10k, and 4k constraints respectively, representing 0.0004\%, 0.0004\%, and 0.002\% of the total training set constraints.

To test different cluster orientations, we reconstructed the original datasets into two artificial versions: default and personalized. We set target clusters to 2 or 4, facilitating division into distinct semantic orientations. We named these datasets CIFAR10-2, CIFAR100-4, and ImageNet10-2 to reflect the differences of labels from the originals. Tab. \ref{table:1} and Appendix illustrate the details of the adopted datasets. The artificial datasets share the same number of target clusters. The default orientation aligns with deep clustering tendencies, while the personalized orientation is intentionally opposite. By comparing results on these datasets, we evaluate whether our PCL outperforms state-of-the-art deep clustering and semi-supervised methods.

\textbf{Compared Methods.}
To verify the effectiveness of the proposed method, we select some representative and frontier methods for comparison. We compare with the known or SOTA clustering methods as follows:

\begin{itemize}
\item \textbf{K-means}~\cite{MacQueen1967SomeMF}: A classical partition-based clustering algorithm.
\item \textbf{MiCE}~\cite{tsai2020mice}: Combines contrastive learning with latent probability models.
\item \textbf{CC}~\cite{Li2021ContrastiveC}: A leading method for contrastive clustering at instance and cluster levels.
\item \textbf{GCC}~\cite{Zhong2021GraphCC}: Extends positive sample pairs using KNN graph neighbors.
\item \textbf{SPICE}~\cite{niu2022spice}: A three-stage training method with excellent clustering performance.
\end{itemize}

Besides, we also compared with constrained clustering methods:

\begin{itemize}
\item \textbf{DCC}~\cite{Zhang2019AFF}: Extends IDEC~\cite{Guo2017ImprovedDE} with constraints.
\item \textbf{ADC}~\cite{Sun2022ActiveDI}: Constructs contrastive losses using only constraint pairs.
\item \textbf{SDEC}~\cite{Ren2019SemisupervisedDE}: Extends DEC~\cite{Xie2016UnsupervisedDE} with constraints.
\item \textbf{DC-GMM}~\cite{Manduchi2021DeepCG}: A recent semi-supervised method using probabilistic domain knowledge.
\end{itemize}

\textbf{Implementation Details.}
All methods used ResNet34 as the backbone network without modification. Parameters related to deep contrastive clustering followed previous methods~\cite{Li2021ContrastiveC,Zhong2021GraphCC}. 
The batch size was 128, and Adam with an initial learning rate of 1e-5 was used for optimization. All images were resized to $128\times128$. The feature dimensionality $M$ was set to 128. Hyperparameters were consistent across datasets with $\lambda=4$ and $\epsilon=0.2$. Constraints were extended by labeling similar pairs with high confidence after several iterations. For semi-supervised and active constrained methods, 10k pairwise constraints were set on the three datasets. Our method used a training epoch $E=500$ to determine final performance.

\textbf{Evaluation Metrics.}
We adopted four standard clustering metrics to evaluate our method including Normalized Mutual Information (NMI), Accuracy (ACC), F-measures (F) , and Adjusted Rand Index (ARI). These metrics reflect the performance of clustering from different aspects, and higher values indicate better performance. 

\subsection{Results}
\label{sec:4-2}
\textbf{Clustering Performance.}
The clustering performance of PCL and comparison methods on four datasets with two cluster orientations are presented in Tables \ref{table:2-1} and \ref{table:2-2}. More compared methods can be found at our Appendix.

Key observations include: \romannumeral1) In the default orientation, both unsupervised and constrained methods perform well, with some unsupervised methods even surpassing constrained ones. Our method performs comparably, with slight gaps in some metrics. \romannumeral2) In the personalized orientation, unsupervised methods perform poorly, and existing constrained methods fail to achieve the desired orientation. However, PCL excels in the personalized orientation, outperforming other methods.

These results confirm our expectations. Unsupervised methods struggle across all orientations, while existing constrained methods cannot adequately control cluster orientation. In contrast, PCL performs well across orientations, effectively identifying useful samples to learn cluster orientations and addressing personalized clustering challenges. Notably, the DC-GMM method is highly dependent on feature extraction results. For fairness, we used the original trainable VGG16 without a pre-trained network when comparing DC-GMM.

Overall, PCL successfully clusters along the desired orientation, surpassing state-of-the-art methods.

\begin{figure}[t]
\centering
\includegraphics[width=1\columnwidth]{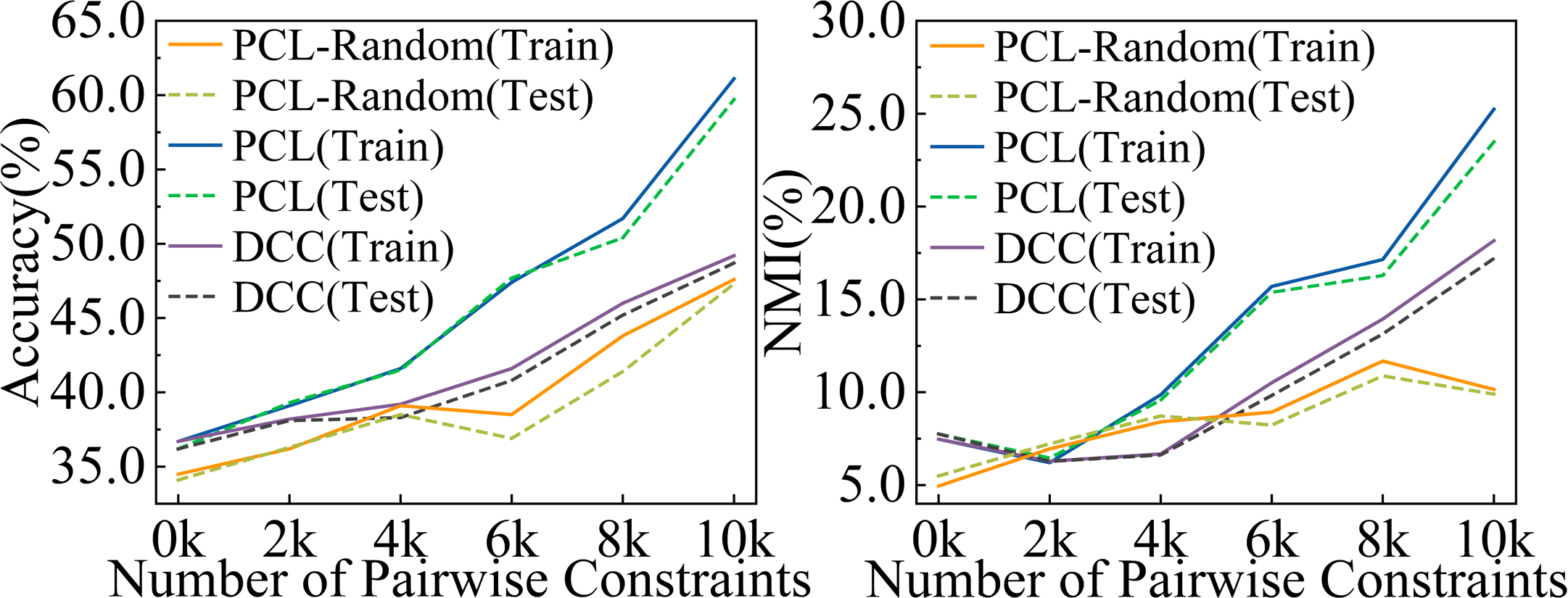} 
\caption{Clustering accuracy and NMI on train sets and test sets for different number of pairwise constraints.}
\label{fig4}
\end{figure}
\textbf{Query Efficiency.}
To evaluate the effectiveness of our method and query strategy, we compared it against a random strategy and DCC, as depicted in Fig. \ref{fig4}. We conducted experiments with varying numbers of constraints on the personalized cluster orientation of the CIFAR100-4 dataset. To further demonstrate that our method effectively learns task-related features, we differentiated between the training and test sets, applying constraints exclusively to the training set.

Key observations include: \romannumeral1) As the number of constraints increases, clustering performance improves. However, with the same number of constraints, our method significantly outperforms the other two methods. \romannumeral2) Our method demonstrates comparable performance on both the test and training sets under identical conditions.

These observations indicate that our method and query strategy excel in performing task-based clustering, focusing on features critical to the task. Additionally, the constraints used (10k) represent only 0.0004\% of the total, highlighting our ability to achieve superior clustering results with minimal constraint information.

Overall, our approach effectively identifies task-related features with fewer queries, addressing personalized clustering challenges efficiently.

\begin{figure}[ht]
\centering
\includegraphics[width=1\columnwidth]{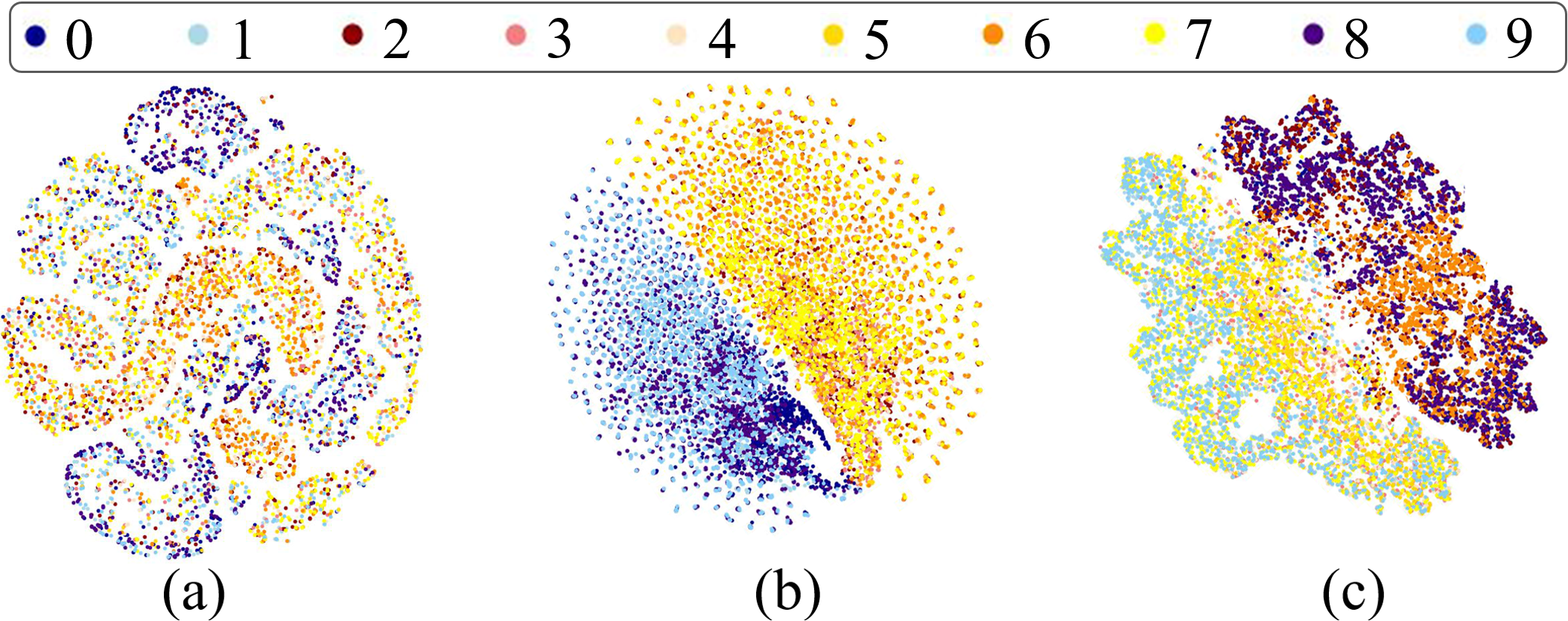} 
\caption{The evolution of features across the training process under two cluster orientations: (a) Initial distribution of samples, (b) Sample distribution after clustering along the default orientation and (c) Sample distribution after clustering along a given personalized orientation. The color of the dots denotes their original class labels of CIFAR-10.}
\label{fig5}
\end{figure}

\begin{table}[ht]
  \centering
    \resizebox{.8\columnwidth}{!}{
    \begin{tabular}{lcccc}
    \toprule
    \multicolumn{1}{l}{Module} & NMI   & ARI  &F  & ACC \\
    \midrule
    CNN only & 0.127 & 0.122 & 0.344 & 0.445  \\
    CNN+Attention  & \textbf{0.234} & \textbf{0.247} & \textbf{0.437} & \textbf{0.597} \\
    \bottomrule
    \end{tabular}%
    }
    \caption{Effect of Cross-Attention Module in personalized clustering on CIFAR100-4.} 
  \label{table:4}%
\end{table}%

\begin{table}[ht]
  \centering
    \resizebox{.8\columnwidth}{!}{
    \begin{tabular}{lcccc}
    \toprule
    \multicolumn{1}{l}{Query strategy} & NMI   & ARI  &F  & ACC \\
    \midrule
    random & 0.172 & 0.170 & 0.380 & 0.487  \\
    $S_{up}$ & 0.196 & 0.211 & 0.411 & 0.569 \\
    $S_{hp}$ & 0.160 & 0.154 & 0.369 & 0.481  \\
    $S_{up}+S_{hs}$ & \textbf{0.234} & \textbf{0.247} & \textbf{0.437} & \textbf{0.597} \\
    \bottomrule
    \end{tabular}%
    }
    \caption{Effect of query strategy in personalized clustering on CIFAR100-4.} 
  \label{table:5}%
\end{table}%

\subsection{Visualization and Ablation Studies.}
\label{sec:4-3}
\textbf{Visualization.} To vividly display the personalized clustering results, we visualize the distribution of samples in CIFAR10-2 after clustering. In Fig. \ref{fig4}, we have the following observations. \romannumeral1) Subfigure (a) depicts that the sample points before clustering is chaos. \romannumeral2) Subfigure (b) depicts that the yellow and orange sample points cluster together, while in subfigure (c) the dark blue and dark orange dots are clustered together. The above facts show that PCL can focus on the relationship between sample representations on the targeted orientation rather than any other orientations.

\textbf{Effect of Cross-Attention Module.} We conducted an ablation analysis by removing the cross-attention module. Results along the personalized cluster orientation of CIFAR100-4 are shown in Table \ref{table:4}. The highest clustering performance was achieved with the cross-attention module, highlighting its necessity for extracting task-specific features.

\textbf{Effect of joint query strategy.} We performed an ablation analysis by removing components of the query strategy. Results along the personalized cluster orientation of CIFAR100-4 are shown in Table \ref{table:5}. Personalized clustering achieved better results using our query strategy.

\section{Conclusion}
\label{sec:5}
In this study, we attempt to tackle the problem of personalized clustering by interacting with users and then propose a clustering model named PCL. In our work, we designed a kind of active query strategy to identify sample pairs that are informative for targeted representation learning. To further enhance the learning process, we propose a constrained clustering loss and leverage attention mechanism, maximizing the power of constraints in guiding the desired cluster orientation. We have theoretically verified the effectiveness of our query strategy and loss function, and extensive experiments have validated our findings, simultaneously.

\newpage
\section*{Acknowledgements}
This work is supported by the National Key Research \& Develop Plan (2023YFB4503600), National Natural Science Foundation of China (U23A20299, U24B20144, 62172424, 62276270, 62322214). It is also partially supported by the Opening Fund of Hebei Key Laboratory of Machine Learning and Computational Intelligence. We sincerely appreciate the valuable and insightful feedback provided by all reviewers.
\bibliography{aaai25}

\begin{thebibliography}{43}
\providecommand{\natexlab}[1]{#1}

\bibitem[{Angell et~al.(2022)Angell, Monath, Yadav, and McCallum}]{Angell2022InteractiveCC}
Angell, R.; Monath, N.; Yadav, N.; and McCallum, A. 2022.
\newblock Interactive Correlation Clustering with Existential Cluster Constraints.
\newblock In \emph{ICML}.

\bibitem[{Ashtiani, Kushagra, and Ben-David(2016)}]{Ashtiani2016ClusteringWS}
Ashtiani, H.; Kushagra, S.; and Ben-David, S. 2016.
\newblock Clustering with Same-Cluster Queries.
\newblock In \emph{Advances in Neural Information Processing Systems}, volume~29. Curran Associates, Inc.

\bibitem[{Bai, Liang, and Cao(2020)}]{bai2020semi}
Bai, L.; Liang, J.; and Cao, F. 2020.
\newblock Semi-supervised clustering with constraints of different types from multiple information sources.
\newblock \emph{IEEE Transactions on pattern analysis and machine intelligence}, 43(9): 3247--3258.

\bibitem[{Basu, Banerjee, and Mooney(2002)}]{Basu2002SemisupervisedCB}
Basu, S.; Banerjee, A.; and Mooney, R.~J. 2002.
\newblock Semi-supervised Clustering by Seeding.
\newblock In \emph{International Conference on Machine Learning}.

\bibitem[{Bilenko, Basu, and Mooney(2004)}]{bilenko2004integrating}
Bilenko, M.; Basu, S.; and Mooney, R.~J. 2004.
\newblock Integrating constraints and metric learning in semi-supervised clustering.
\newblock In \emph{Proceedings of the twenty-first international conference on Machine learning}, 11.

\bibitem[{Biswas and Jacobs(2014)}]{Biswas2014ActiveIC}
Biswas, A.; and Jacobs, D.~W. 2014.
\newblock Active Image Clustering with Pairwise Constraints from Humans.
\newblock \emph{International Journal of Computer Vision}, 108: 133--147.

\bibitem[{Caron et~al.(2018)Caron, Bojanowski, Joulin, and Douze}]{Caron2018DeepCF}
Caron, M.; Bojanowski, P.; Joulin, A.; and Douze, M. 2018.
\newblock Deep Clustering for Unsupervised Learning of Visual Features.
\newblock In \emph{ECCV}.

\bibitem[{Chien, Zhou, and Li(2019)}]{Chien2019HS2AL}
Chien, E.; Zhou, H.; and Li, P. 2019.
\newblock HS2: Active learning over hypergraphs with pointwise and pairwise queries.
\newblock In \emph{AISTATS}.

\bibitem[{Dasgupta and Ng(2010)}]{dasgupta2010clustering}
Dasgupta, S.; and Ng, V. 2010.
\newblock Which clustering do you want? inducing your ideal clustering with minimal feedback.
\newblock \emph{Journal of Artificial Intelligence Research}, 39: 581--632.

\bibitem[{Deng et~al.(2009)Deng, Dong, Socher, Li, Li, and Fei-Fei}]{deng2009imagenet}
Deng, J.; Dong, W.; Socher, R.; Li, L.-J.; Li, K.; and Fei-Fei, L. 2009.
\newblock Imagenet: A large-scale hierarchical image database.
\newblock In \emph{2009 IEEE conference on computer vision and pattern recognition}, 248--255. Ieee.

\bibitem[{Gan et~al.(2023)Gan, Zhao, Kang, Shang, Chen, and Li}]{gan2023superclass}
Gan, Z.; Zhao, S.; Kang, J.; Shang, L.; Chen, H.; and Li, C. 2023.
\newblock Superclass Learning with Representation Enhancement.
\newblock In \emph{Proceedings of the IEEE/CVF Conference on Computer Vision and Pattern Recognition}, 24060--24069.

\bibitem[{Guo et~al.(2017)Guo, Gao, Liu, and Yin}]{Guo2017ImprovedDE}
Guo, X.; Gao, L.; Liu, X.; and Yin, J. 2017.
\newblock Improved Deep Embedded Clustering with Local Structure Preservation.
\newblock In \emph{International Joint Conference on Artificial Intelligence}.

\bibitem[{Huang, Gong, and Zhu(2020)}]{Huang2020DeepSC}
Huang, J.; Gong, S.; and Zhu, X. 2020.
\newblock Deep Semantic Clustering by Partition Confidence Maximisation.
\newblock \emph{2020 IEEE/CVF Conference on Computer Vision and Pattern Recognition (CVPR)}, 8846--8855.

\bibitem[{Jain, Nandakumar, and Ross(2005)}]{jain2005score}
Jain, A.; Nandakumar, K.; and Ross, A. 2005.
\newblock Score normalization in multimodal biometric systems.
\newblock \emph{Pattern recognition}, 38(12): 2270--2285.

\bibitem[{Kingma and Welling(2014)}]{Kingma2014AutoEncodingVB}
Kingma, D.~P.; and Welling, M. 2014.
\newblock Auto-Encoding Variational Bayes.
\newblock \emph{CoRR}, abs/1312.6114.

\bibitem[{Krizhevsky(2009)}]{Krizhevsky2009LearningML}
Krizhevsky, A. 2009.
\newblock Learning Multiple Layers of Features from Tiny Images.
\newblock Technical report, University of Toronto, Toronto.

\bibitem[{Li et~al.(2021)Li, Hu, Liu, Peng, Zhou, and Peng}]{Li2021ContrastiveC}
Li, Y.; Hu, P.; Liu, Z.; Peng, D.; Zhou, J.~T.; and Peng, X. 2021.
\newblock Contrastive Clustering.
\newblock \emph{Proceedings of the AAAI Conference on Artificial Intelligence}, 35(10): 8547--8555.

\bibitem[{Li et~al.(2023)Li, Hu, Peng, Lv, Fan, and Peng}]{li2023image}
Li, Y.; Hu, P.; Peng, D.; Lv, J.; Fan, J.; and Peng, X. 2023.
\newblock Image clustering with external guidance.
\newblock \emph{arXiv preprint arXiv:2310.11989}.

\bibitem[{Li et~al.(2022)Li, Yang, Peng, Li, Huang, and Peng}]{li2022twin}
Li, Y.; Yang, M.; Peng, D.; Li, T.; Huang, J.; and Peng, X. 2022.
\newblock Twin contrastive learning for online clustering.
\newblock \emph{International Journal of Computer Vision}, 130(9): 2205--2221.

\bibitem[{Lin, Xu, and Zhang(2019)}]{Lin2019DiscoveringNI}
Lin, T.-E.; Xu, H.; and Zhang, H. 2019.
\newblock Discovering New Intents via Constrained Deep Adaptive Clustering with Cluster Refinement.
\newblock In \emph{AAAI Conference on Artificial Intelligence}.

\bibitem[{Liu et~al.(2022)Liu, Tu, Zhou, Liu, Song, Yang, and Zhu}]{Liu2022DeepGC}
Liu, Y.; Tu, W.; Zhou, S.; Liu, X.; Song, L.; Yang, X.; and Zhu, E. 2022.
\newblock Deep Graph Clustering via Dual Correlation Reduction.
\newblock In \emph{AAAI}.

\bibitem[{MacQueen(1967)}]{MacQueen1967SomeMF}
MacQueen, J. 1967.
\newblock Some methods for classification and analysis of multivariate observations.
\newblock In \emph{Fifth Berkeley Symposium on Mathematical Statistics and Probability}, 281–297.

\bibitem[{Manduchi et~al.(2021)Manduchi, Chin-Cheong, Michel, Wellmann, and Vogt}]{Manduchi2021DeepCG}
Manduchi, L.; Chin-Cheong, K.; Michel, H.; Wellmann, S.; and Vogt, J.~E. 2021.
\newblock Deep Conditional Gaussian Mixture Model for Constrained Clustering.
\newblock In \emph{NeurIPS}.

\bibitem[{Niu, Shan, and Wang(2022)}]{niu2022spice}
Niu, C.; Shan, H.; and Wang, G. 2022.
\newblock Spice: Semantic pseudo-labeling for image clustering.
\newblock \emph{IEEE Transactions on Image Processing}, 31: 7264--7278.

\bibitem[{Oord, Li, and Vinyals(2018)}]{oord2018representation}
Oord, A. v.~d.; Li, Y.; and Vinyals, O. 2018.
\newblock Representation learning with contrastive predictive coding.
\newblock \emph{arXiv preprint arXiv:1807.03748}.

\bibitem[{Pei, Liu, and Fern(2015)}]{Pei2015BayesianAC}
Pei, Y.; Liu, L.-P.; and Fern, X.~Z. 2015.
\newblock Bayesian Active Clustering with Pairwise Constraints.
\newblock In \emph{ECML/PKDD}.

\bibitem[{Ren et~al.(2022)Ren, Xiao, Chang, Huang, Li, Chen, and Wang}]{Ren2022ASO}
Ren, P.; Xiao, Y.; Chang, X.; Huang, P.; Li, Z.; Chen, X.; and Wang, X. 2022.
\newblock A Survey of Deep Active Learning.
\newblock \emph{ACM Computing Surveys (CSUR)}, 54: 1--40.

\bibitem[{Ren et~al.(2019)Ren, Hu, Dai, Pan, Hoi, and Xu}]{Ren2019SemisupervisedDE}
Ren, Y.; Hu, K.; Dai, X.; Pan, L.; Hoi, S. C.~H.; and Xu, Z. 2019.
\newblock Semi-supervised deep embedded clustering.
\newblock \emph{Neurocomputing}, 325: 121--130.

\bibitem[{Saunshi et~al.(2019)Saunshi, Plevrakis, Arora, Khodak, and Khandeparkar}]{pmlr-v97-saunshi19a}
Saunshi, N.; Plevrakis, O.; Arora, S.; Khodak, M.; and Khandeparkar, H. 2019.
\newblock A Theoretical Analysis of Contrastive Unsupervised Representation Learning.
\newblock In Chaudhuri, K.; and Salakhutdinov, R., eds., \emph{Proceedings of the 36th International Conference on Machine Learning}, volume~97 of \emph{Proceedings of Machine Learning Research}, 5628--5637. PMLR.

\bibitem[{Sun et~al.(2022)Sun, Zhou, Du, and Li}]{Sun2022ActiveDI}
Sun, B.; Zhou, P.; Du, L.; and Li, X. 2022.
\newblock Active deep image clustering.
\newblock \emph{Knowl. Based Syst.}, 252: 109346.

\bibitem[{Tao, Takagi, and Nakata(2021)}]{TaoTN21}
Tao, Y.; Takagi, K.; and Nakata, K. 2021.
\newblock Clustering-friendly Representation Learning via Instance Discrimination and Feature Decorrelation.
\newblock In \emph{9th International Conference on Learning Representations (ICLR)}.

\bibitem[{Tsai, Li, and Zhu(2020)}]{tsai2020mice}
Tsai, T.~W.; Li, C.; and Zhu, J. 2020.
\newblock Mice: Mixture of contrastive experts for unsupervised image clustering.
\newblock In \emph{International conference on learning representations}.

\bibitem[{van Craenendonck, Dumancic, and Blockeel(2018)}]{Craenendonck2018COBRAAF}
van Craenendonck, T.; Dumancic, S.; and Blockeel, H. 2018.
\newblock COBRA: A Fast and Simple Method for Active Clustering with Pairwise Constraints.
\newblock In \emph{International Joint Conference on Artificial Intelligence}.

\bibitem[{Vincent et~al.(2010)Vincent, Larochelle, Lajoie, Bengio, and Manzagol}]{Vincent2010StackedDA}
Vincent, P.; Larochelle, H.; Lajoie, I.; Bengio, Y.; and Manzagol, P.-A. 2010.
\newblock Stacked Denoising Autoencoders: Learning Useful Representations in a Deep Network with a Local Denoising Criterion.
\newblock \emph{J. Mach. Learn. Res.}, 11: 3371--3408.

\bibitem[{Wagstaff et~al.(2001)Wagstaff, Cardie, Rogers, and Schr{\"o}dl}]{Wagstaff2001ConstrainedKC}
Wagstaff, K.~L.; Cardie, C.; Rogers, S.; and Schr{\"o}dl, S. 2001.
\newblock Constrained K-means Clustering with Background Knowledge.
\newblock In \emph{International Conference on Machine Learning}.

\bibitem[{Wu et~al.(2019)Wu, Long, Wang, Qian, Li, Lin, and Zha}]{Wu2019DeepCC}
Wu, J.; Long, K.; Wang, F.; Qian, C.; Li, C.; Lin, Z.; and Zha, H. 2019.
\newblock Deep Comprehensive Correlation Mining for Image Clustering.
\newblock \emph{2019 IEEE/CVF International Conference on Computer Vision (ICCV)}, 8149--8158.

\bibitem[{Xie, Girshick, and Farhadi(2016)}]{Xie2016UnsupervisedDE}
Xie, J.; Girshick, R.; and Farhadi, A. 2016.
\newblock Unsupervised Deep Embedding for Clustering Analysis.
\newblock In \emph{Proceedings of The 33rd International Conference on Machine Learning}, volume~48 of \emph{Proceedings of Machine Learning Research}, 478--487. PMLR.

\bibitem[{Yang et~al.(2013)Yang, Rutayisire, Lin, Li, and Teng}]{Yang2013AnIC}
Yang, Y.; Rutayisire, T.; Lin, C.; Li, T.; and Teng, F. 2013.
\newblock An Improved Cop-Kmeans Clustering for Solving Constraint Violation Based on MapReduce Framework.
\newblock \emph{Fundam. Informaticae}, 126: 301--318.

\bibitem[{Yang et~al.(2012)Yang, Tan, Li, and Ruan}]{Yang2012ConsensusCB}
Yang, Y.; Tan, W.; Li, T.; and Ruan, D. 2012.
\newblock Consensus clustering based on constrained self-organizing map and improved Cop-Kmeans ensemble in intelligent decision support systems.
\newblock \emph{Knowl. Based Syst.}, 32: 101--115.

\bibitem[{Zhang, Basu, and Davidson(2019)}]{Zhang2019AFF}
Zhang, H.; Basu, S.; and Davidson, I. 2019.
\newblock A Framework for Deep Constrained Clustering - Algorithms and Advances.
\newblock In \emph{ECML/PKDD}.

\bibitem[{Zheng and Li(2011)}]{Zheng2011SemisupervisedHC}
Zheng, L.; and Li, T. 2011.
\newblock Semi-supervised Hierarchical Clustering.
\newblock \emph{2011 IEEE 11th International Conference on Data Mining}, 982--991.

\bibitem[{Zhong et~al.(2021)Zhong, Wu, Chen, Huang, Deng, Nie, Lin, and Hua}]{Zhong2021GraphCC}
Zhong, H.; Wu, J.; Chen, C.; Huang, J.; Deng, M.; Nie, L.; Lin, Z.; and Hua, X. 2021.
\newblock Graph Contrastive Clustering.
\newblock \emph{2021 IEEE/CVF International Conference on Computer Vision (ICCV)}, 9204--9213.

\bibitem[{Zhuang and Moulin(2023)}]{zhuang2023deep}
Zhuang, F.; and Moulin, P. 2023.
\newblock Deep semi-supervised metric learning with mixed label propagation.
\newblock In \emph{Proceedings of the IEEE/CVF conference on computer vision and pattern recognition}, 3429--3438.

\end{thebibliography}
\clearpage

\appendix

\section{Algorithm}
\label{sec:algorithm}

\begin{algorithm}[ht]
\caption{Personalized Clustering}
\textbf{Input}: Training dataset $\mathcal X$; Cluster number $K$; Training epochs $E$; Batch size $N$.\\
\textbf{Parameter}: Initial query quantity $Q_1$, Query times $Q$; Temperature parameter $\tau$; Hyper-parameters $\lambda,\epsilon$.\\
\textbf{Output}: The clustering result $\mathcal C$.

\begin{algorithmic}[1] 
\STATE Randomly query $Q_1$ constraints.
\FOR{$epoch=1$ to $E$}
\FOR{a instanced mini-batch $\{\textbf{x}_i\}_{i=1}^N$}
\STATE Generating augmented data $\{\textbf x_1,..,\textbf x_{2N}\}$ and construct indicator matrix $\indicator{}$;
\STATE Utilize neural network to extract feature matrix;
\STATE Project feature onto the feature space to get $\mathbf Z$;
\STATE Project feature onto the cluster space to get $\mathbf P$;
\STATE Select the instance pair with $\max_{(\textbf x_i,\textbf x_j)}\mathcal S(\textbf x_i,\textbf x_j)$ in Eq.8;
\STATE The oracle indicates the instance pair, and it is recorded with indicator matrix $\indicator{}^+$ and $\indicator{}^-$;
\STATE Calculate the constrained contrastive loss $\mathcal L$ through Eq.1--5;
\STATE Update the network by minimizing $\mathcal L$;
\ENDFOR
\ENDFOR
\STATE Calculate $\mathcal C=\arg\max(\textbf P)$
\STATE \textbf{return} $\mathcal C$
\end{algorithmic}
\end{algorithm}

\begin{figure}[ht]
\centering
\includegraphics[width=1\columnwidth]{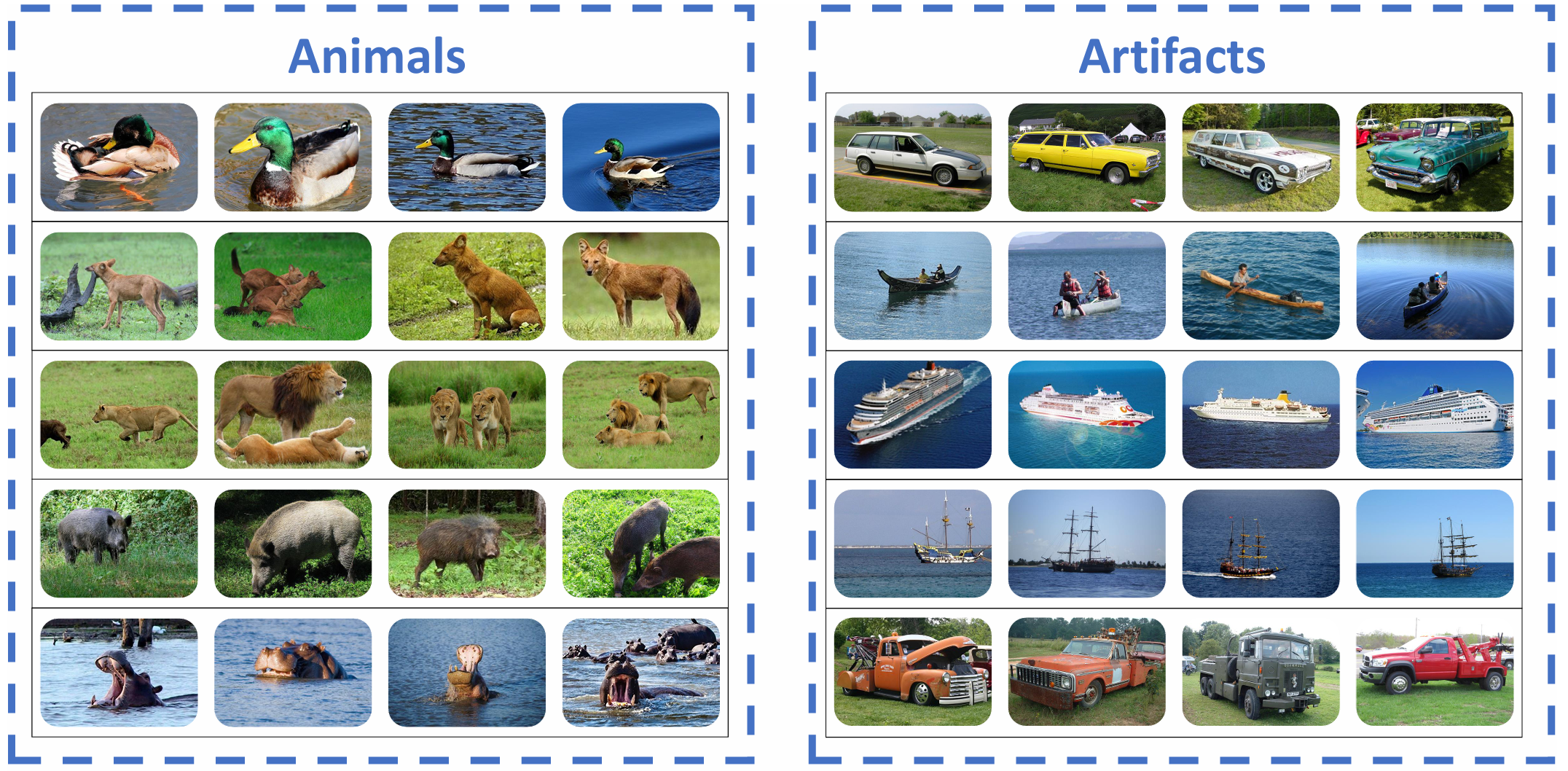}
\caption{The default orientation following the tendency of unsupervised clustering.}
\label{fig6}
\end{figure}

\begin{figure}[ht]
\centering
\includegraphics[width=1\columnwidth]{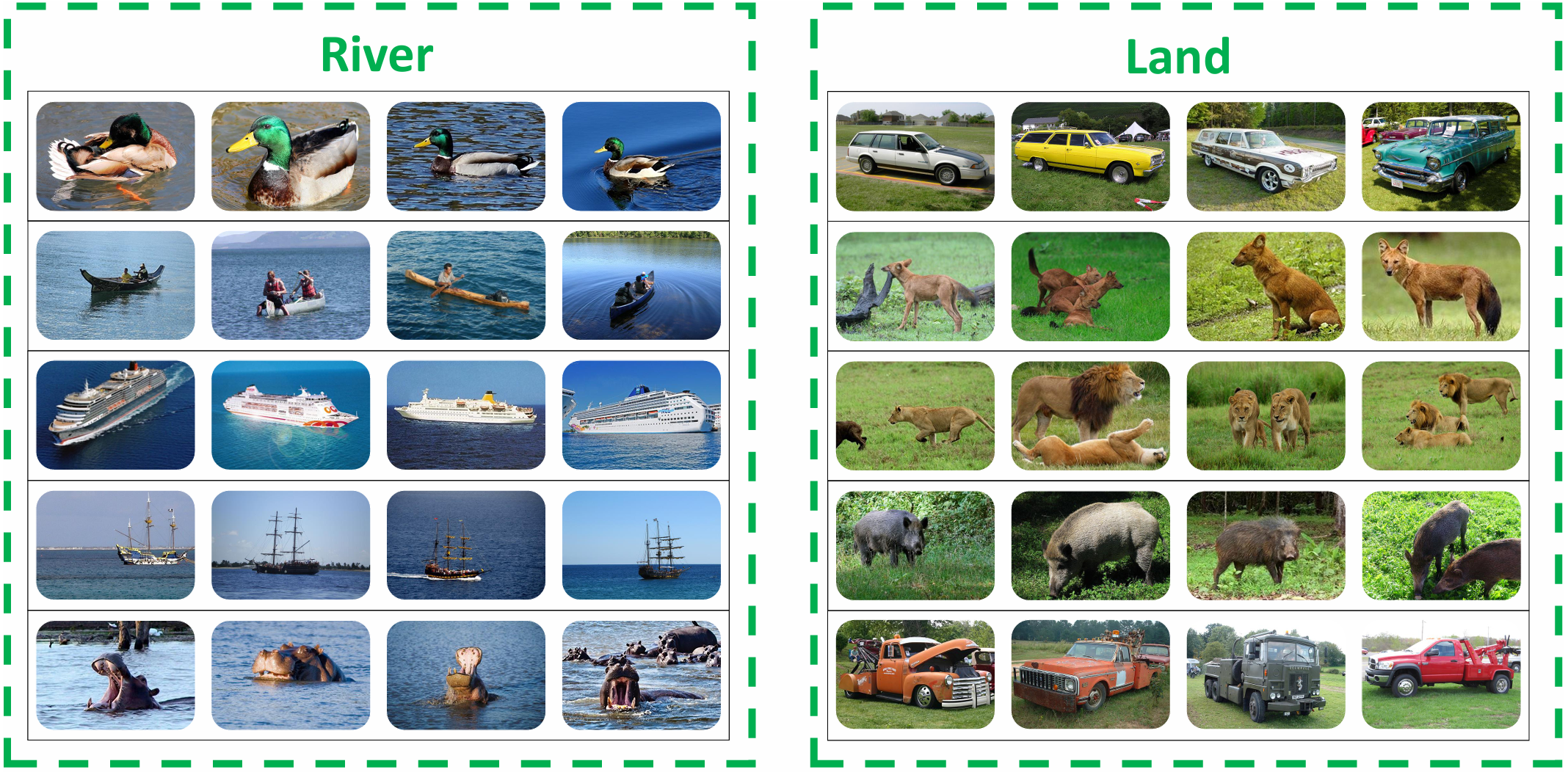}
\caption{The personalized orientation designed artificially.}
\label{fig7}
\end{figure}

\section{Datasets Implementation details}
\label{sec:dataset}
To simulate the diversity of image data clustering orientation, we set the target clusters to 2 or 4 and relabel the datasets. It is expensive to relabel all images to create another clustering demand, while it is easy to divide the 10 clusters into 2 clusters in two ways. Thus, we redesigned the target clusters into default and personalization orientations. The default orientation completely follows the tendency of deep clustering. The personalized target orientation is designed opposite to the default one. It is another orientation that conforms to reality. Specifically, CIFAR-10 and ImageNet-10 grouping according to the background. CIFAR-100 is divided into four categories: large animals, small animals, plants, and artifacts. The specific Settings that map the original classes to our target cluster are shown in the Tab. \ref{tab:6}--\ref{tab:11}. For every original datasets, we construct two clustering tasks by designing two clustering orientations: default and personalized. Fig.\ref{fig6} and Fig.\ref{fig7} show an example of two clustering orientations from the same original dataset.

\begin{minipage}{\columnwidth}
\begin{minipage}[ht]{0.45\columnwidth}
\makeatletter\def\@captype{table}
    \begin{tabular}{cc}
    \toprule
    Original & Target \\
    \midrule
    airplane & \multirow{4}[2]{*}{Artifacts} \\
    automobile &  \\
    ship  &  \\
    truck &  \\
    \midrule
    bird  & \multirow{6}[2]{*}{Animals} \\
    cat   &  \\
    deer  &  \\
    dog   &  \\
    frog  &  \\
    horse &  \\
    \bottomrule
    \end{tabular}%
\caption{Default orientation in CIFAR-10.}
\label{tab:6}
\end{minipage}
\begin{minipage}[ht]{0.45\columnwidth}
\makeatletter\def\@captype{table}
    \begin{tabular}{cc}
    \toprule
    \multicolumn{1}{c}{Original} & Target \\
    \midrule
    airplane & \multirow{4}[2]{*}{Others} \\
    bird  &  \\
    frog  &  \\
    ship  &  \\
    \midrule
    automobile & \multirow{6}[2]{*}{Land} \\
    cat   &  \\
    deer  &  \\
    dog   &  \\
    horse &  \\
    truck &  \\
    \bottomrule
    \end{tabular}%
\caption{Personalize orientation in CIFAR-10.}
\label{tab:7}
\end{minipage}
\end{minipage}

\begin{minipage}{\columnwidth}
\begin{minipage}[ht]{0.45\columnwidth}
\makeatletter\def\@captype{table}
    \begin{tabular}{cc}
    \toprule
    Original & Target \\
    \midrule
    drake & \multirow{5}[2]{*}{Animals} \\
    dhole &  \\
    lion  &  \\
    boar  &  \\
    hippo &  \\
    \midrule
    check & \multirow{5}[2]{*}{Artifacts} \\
    canoe &  \\
    liner &  \\
    pirate &  \\
    truck &  \\
    \bottomrule
    \end{tabular}%
\caption{Default orientation in ImageNet-10.}
\label{tab:10}
\end{minipage}
\begin{minipage}[ht]{0.45\columnwidth}
\makeatletter\def\@captype{table}
    \begin{tabular}{cc}
    \toprule
    Original & Target \\
    \midrule
    drake & \multirow{5}[2]{*}{River} \\
    hippo &  \\
    canoe &  \\
    liner &  \\
    pirate &  \\
    \midrule
    dhole & \multirow{5}[2]{*}{Land} \\
    lion  &  \\
    boar  &  \\
    check &  \\
    truck &  \\
    \bottomrule
    \end{tabular}%
\caption{Personalize orientation in ImageNet-10.}
\label{tab:11}
\end{minipage}
\end{minipage}

\begin{table*}[t]
  \centering
    \begin{tabular}{p{16cm}c}
    \toprule
    Original & Target \\
    \midrule
    apple, beetle, bicycle, bowl, clock, cockroach, crab, cup, dinosaur, hamster, keyboard, lawn\_mower, orange, orchid, pear, plate, poppy, rose, spider, sunflower, sweet\_pepper, telephone, tulip, worm & defalut 0 \\
    \midrule
    beaver, bee, butterfly, camel, caterpillar, cattle, crocodile, elephant, forest, fox, kangaroo, leopard, lion, lizard, lobster, maple\_tree, motorcycle, mouse, mushroom, oak\_tree, palm\_tree, pine\_tree, porcupine, shrew, snail, snake, squirrel, tiger, willow\_tree & defalut 1 \\
    \midrule
    bed, bottle, bridge, bus, can, castle, chair, couch, house, lamp, mountain, pickup\_truck, plain, road, rocket, sea, skyscraper, streetcar, table, tank, television, tractor, train, wardrobe & defalut 2 \\
    \midrule
    aquarium\_fish, baby, bear, boy, chimpanzee, cloud, dolphin, flatfish, girl, man, otter, possum, rabbit, raccoon, ray, seal, shark, skunk, trout, turtle, whale, wolf, woman & defalut 3 \\
    \bottomrule
    \end{tabular}%
  \caption{Default orientation in CIFAR-100.}
  \label{tab:8}%
\end{table*}%

\begin{table*}[t]
  \centering
    \begin{tabular}{p{15.3cm}c}
    \toprule
    Original & Target \\
    \midrule
    baby, bear, beaver, boy, camel, cattle, chimpanzee, dolphin, elephant, fox, girl, kangaroo, leopard, lion, man, otter, porcupine, possum, raccoon, seal, skunk, tiger, whale, wolf, woman & macroanimal \\
    \midrule
    aquarium\_fish, bee, beetle, butterfly, caterpillar, cockroach, crab, crocodile, dinosaur, flatfish, hamster, lizard, lobster, mouse, rabbit, shark, shrew, snail, snake, spider, squirrel, trout, turtle, worm & Small animal \\
    \midrule
    apple, cloud, forest, maple\_tree, mountain, mushroom, oak\_tree, orange, orchid, palm\_tree, pear, pine\_tree, plain, poppy, ray, rose, sea, sunflower, sweet\_pepper, tulip, willow\_tree & plant \\
    \midrule
    bed, bicycle, bottle, bowl, bridge, bus, can, castle, chair, clock, couch, cup, house, keyboard, lamp, lawn\_mower, motorcycle, pickup\_truck, plate, road, rocket, skyscraper, streetcar, table, tank, telephone, television, tractor, train, wardrobe & artifact \\
    \bottomrule
    \end{tabular}%
  \caption{Personalize orientation in CIFAR-100.}
  \label{tab:9}%
\end{table*}%

\section{Theorem}
\subsection{Proof of Theorem 3.1 and Corollary 3.1}
First we note that $l$ is bounded by $B$ ($B>0$), applying Hoeffding's inequality, 
with probability $1-\delta$, for all $f$, we have:
\begin{equation}
    \hat{\mathcal{L}}_{Q}(f) - \mathcal{L}_{Q}(f) \leq 3B\sqrt{\frac{\log\frac{1}{\delta}}{2N}}.
\end{equation}

\subsubsection{Theorem A.1}\textit{ Let $l: \mathbb{R}^M\rightarrow \mathbb{R}$ be $\eta$-Lipschitz and bounded by $B$. Then with probability at least $1-\delta$ over the training set $\mathcal{S} = \{\mathbf {x}_1,...,\mathbf{x}_{2N}\}$, we have: }
\begin{equation}
    |\mathcal{L}_{Q}(\hat{f}_Q) - \mathcal{L}_{Q}(f^*_Q)| \leq c\frac{\eta \mathcal{R}_{\mathcal{S}}(\mathcal{F}) \sqrt{M}}{N} + 3B\sqrt{\frac{\log\frac{1}{\delta}}{2N}}.
\end{equation}

\textit{where $\mathcal{R}_{\mathcal{S}}(\mathcal{F}) =  \mathbb{E}_{\sigma}\left[\sup_{f\in \mathcal{F}}\frac{1}{2N}\sum_{i=1}^{2N}\sigma_i f(\mathbf{x}_i)\right]$ is the Rademacher complexity of $\mathcal{F}$ with respect to $\mathcal{S}$, and $\sigma = (\sigma_1,...,\sigma_{2N})$ is a Rademacher sequence, $c$ is constant.}

Theorem A.1 has been proved by Saunshi et al. in 2019. 

Note that: 
\begin{equation}
\begin{aligned}
    & | \hat{\mathcal{L}}_Q(\hat{f}) - \mathcal{L}_Q(f^*)| \\
    & \leq | \hat{\mathcal{L}}_Q(\hat{f}) - \mathcal{L}_Q(\hat{f})| + | \mathcal{L}_Q(\hat{f}) - \mathcal{L}_Q(f^*)|
\end{aligned}
\end{equation}

Combining (1),(3) and Theorem A.1, we use the union bound get the result of Theorem 3.1 :
\subsubsection{Theorem 3.1} \textit{With probability at least $1-\delta$},
\begin{equation}
\label{eq:17}
    | \mathcal{\hat{L}}_{Q}(\hat{f}_{Q}) - \mathcal{L}_{Q}(f^{*}_{Q}) | \leq \mathcal{O}(\frac{\eta \mathcal{R}_{\mathcal{S}}(\mathcal{F})}{N} + \sqrt{\frac{\log\frac{1}{\delta}}{N}}).
\end{equation}
$\hfill\qedsymbol$
\subsection{Proof of Theorem 3.2 and Theorem 3.3}
When $q \ll N$, the normalized term $\mu_i$ and the matrix $\mathbf{W}$ are approximately constant. Suppose we query the pair $(\mathbf x_i, \mathbf x_j)$ while the feature extractor $f$ is fixed, 
we can observe that $\hat{\mathcal{L}}_{q}(f) \leq \hat{\mathcal{L}}_{q+1}(f)$ 
because either $\omega_{ij}$ or $\indicator{}_{ij}^{-}$ is updated. 
Due to the definitions, for all $f$, we have:
\begin{equation}
        \hat{\mathcal{L}}_{q}(\hat{f}) \leq \hat{\mathcal{L}}_{q+1}(f)
\end{equation}

The left side of (2) is constant, so we derive $\hat{\mathcal{L}}_{q}(\hat{f}_{q}) \leq \hat{\mathcal{L}}_{q+1}(\hat{f}_{q+1})\leq \hat{\mathcal{L}}_{q+2}(\hat{f}_{q+2})\leq\dots\leq\hat{\mathcal{L}}_{Q}(\hat{f}_{Q})$. So we have:
\begin{equation}
    |\mathcal{\hat{L}}_{q}(\hat{f}_{q}) - \mathcal{\hat{L}}_{Q}(\hat{f}_{Q})| \geq |\mathcal{\hat{L}}_{q+1}(\hat{f}_{q+1}) - \mathcal{\hat{L}}_{Q}(\hat{f}_{Q})|
\end{equation}

Then we prove Theorem 3.3.

For the cannot-link case, if we query $(\mathbf x_i^{'}, \mathbf x_j^{'})$, we have:
\begin{equation*}
    \begin{aligned}
        & \quad |\mathcal{\hat{L}}_{q}(f) - \mathcal{\hat{L}}_{q+1}(f)| \\
        &= \frac{1}{2N} \sum_{i=1}^{2N}\sum_{j=1}^{2N}\frac{\omega_{ij} }{\sum_{k=1}^{2N}\omega_{ik}} \cdot \Bigg(  -\log \mu_i \cdot \\
        &  \quad  \Big(   \sum\limits_{k=1}\limits^{2N}(1+\indicator{}_{ik}^-)\exp(s(f(\mathbf{x}_{i}),f(\mathbf{x}_{j}))/\tau) \Big) + \log \mu_i^{'} \cdot \\
        &    \quad  \Big(\sum\limits_{k=1}\limits^{2N}(1+\indicator{}_{ik}^-)\exp(s(f(\mathbf{x}_{i}),f(\mathbf{x}_{j}))/\tau) +\\
        & \quad \exp(s(f(\mathbf{x}_{i}^{'}),f(\mathbf{x}_{j}^{'}))/\tau)   \Big) \Bigg) \\
        &= c_1 + c_2^{'} \cdot \log \left( 1+ \frac{   \exp(s(f(\mathbf{x}_{i}^{'}),f(\mathbf{x}_{j}^{'}))/\tau)  } {c_3  }  \right) \\
        & \approx c_1 + c_2^{''} \cdot  \exp(s(f(\mathbf{x}_{i}^{'}),f(\mathbf{x}_{j}^{'}))/\tau) \\
        &\propto \exp \left(\mathcal{S}_{up}(\mathbf x_i^{'}, \mathbf x_j^{'}) \right).   
    \end{aligned} 
\end{equation*}
where $c_1, c_2, c_2^{'},c_2^{''}, c_3 $ are constants, and $\mathcal{S}_{up}(\cdot,\cdot)$ is the uncertainty score function.

For the must-link case, if we query $(\mathbf x_i^{'}, \mathbf x_j^{'})$, then $\omega_{i^{'}j^{'}}$ will jump from $0$ to $\lambda N$. we have:
\begin{equation*}
    \begin{aligned}
        & |\mathcal{\hat{L}}_{q}(f) - \mathcal{\hat{L}}_{q+1}(f)|\\
        &= \frac{1}{2N} \sum_{i=1}^{2N}\sum_{j=1}^{2N}\frac{\omega_{ij}  \left(  
        -\log \frac{\exp(s(f(\mathbf{x}_{i}),f(\mathbf{x}_{j}))/\tau)}
        {  \mu_i\sum\limits_{k=1}\limits^{2N}(1+\indicator{}_{ik}^-) \exp(s(f(\mathbf{x}_{i}),f(\mathbf{x}_{j}))/\tau)   }
        \right)}
        {\sum_{k=1}^{2N}\omega_{ik}} - \\
        & \Bigg( \frac{1}{2N} \sum_{i=1}^{2N}\sum_{j=1}^{2N}\frac{\omega_{ij}  \left(  
        -\log \frac{\exp(s(f(\mathbf{x}_{i}),f(\mathbf{x}_{j}))/\tau)}
        {  \mu_i\sum\limits_{k=1}\limits^{2N}(1+\indicator{}_{ik}^-) \exp(s(f(\mathbf{x}_{i}),f(\mathbf{x}_{j}))/\tau)   }
        \right)}
        {\sum_{k=1}^{2N}\omega_{ik}} + \\
        & \frac{\lambda N}{\sum_{k=1}^{2N}\omega_{ik} + \lambda N} \log \frac{\exp(s(f(\mathbf{x}_{i}^{'}),f(\mathbf{x}_{j}^{'}))/\tau) }{\mu_i^{'}\sum\limits_{k=1}\limits^{2N}(1+\indicator{}_{i^{'} k}^-) \exp(s(f(\mathbf{x}_{i}^{'}),f(\mathbf{x}_{j}))/\tau)}\Bigg)\\
        &\approx c_4 + c_5 \cdot \log \frac{\exp(s(f(\mathbf{x}_{i}^{'}),f(\mathbf{x}_{j}^{'}))/\tau) }{c_6} \\
        & \propto \mathcal{S}_{up}(\mathbf x_i^{'}, \mathbf x_j^{'})
    \end{aligned}
\end{equation*}
All proofs are completed. 
$\hfill\qedsymbol$

\end{document}